\pdfoutput=1
\documentclass[11pt]{article}

\usepackage[hyperref]{emnlp2021}

\usepackage{times}
\usepackage{latexsym}

\usepackage[T1]{fontenc}
\usepackage[utf8]{inputenc}

\usepackage{microtype}
\usepackage{tabularx}
\usepackage{booktabs}
\usepackage{multicol}
\usepackage{multirow}
\usepackage{amsmath}
\usepackage{amsfonts}
\usepackage[algoruled,linesnumbered,vlined]{algorithm2e}
\usepackage{graphicx}

\title{Sent2Span: Span Detection for PICO Extraction in the Biomedical Text without Span Annotations}

\author{Shifeng Liu$^1$, Yifang Sun$^2$, Bing Li$^3$, Wei Wang$^4$, Florence T. Bourgeois$^{5,6}$, Adam G. Dunn$^{5,1}$\\
	\small$^1$The University of Sydney, Australia	
	\small$^2$Northeastern University, China
	\small$^3$IHPC, A*STAR, Singapore\\
	\small$^4$Hong Kong University of Science and Technology (Guangzhou), China\\
	\small$^5$Computational Health Informatics Program, Boston Children’s Hospital, Boston, MA, United States\\
	\small$^6$Department of Pediatrics, Harvard Medical School, Boston, MA, United States \\
	{\small \tt \{shifeng.liu,adam.dunn\}@sydney.edu.au, sunyifang@cse.neu.edu.cn, weiwcs@ust.hk}\\
	{\small \tt li\_bing@ihpc.a-star.edu.sg, florence.bourgeois@childrens.harvard.edu}
}

\begin{document}
\maketitle
\begin{abstract}
The rapid growth in published clinical trials makes it difficult to maintain up-to-date systematic reviews, which require finding all relevant trials. 
This leads to policy and practice decisions based on out-of-date, incomplete, and biased subsets of available clinical evidence. 
Extracting and then normalising Population, Intervention, Comparator, and Outcome~(PICO) information from clinical trial articles may be an effective way to automatically assign trials to systematic reviews and avoid searching and screening—the two most time-consuming systematic review processes.
We propose and test a novel approach to PICO span detection. 
The major difference between our proposed method and previous approaches comes from detecting spans without needing annotated span data and using only crowdsourced sentence-level annotations.
Experiments on two datasets show that PICO span detection results achieve much higher results for recall when compared to fully supervised methods
with PICO sentence detection at least as good as human annotations.
By removing the reliance on expert annotations for span detection, this work could be used in a human-machine pipeline for turning low-quality, crowdsourced, and sentence-level PICO annotations into structured information that can be used to quickly assign trials to relevant systematic reviews.\end{abstract}

\section{Introduction}

Systematic reviews are a critical part of regulatory and clinical decision-making because they are designed to robustly make sense of all available evidence from primary research, 
especially clinical trials, accounting for study design quality and heterogeneity. 
Searching and screening for reports of clinical trials are time-consuming tasks that require specialised expertise but are a necessary component of systematic reviews. 
The rapid rate at which papers were published about COVID-19~\cite{DBLP:journals/bib/WangL21} highlights the need for tools to improve the efficiency of systematic reviews.

A range of methods have been developed to help reduce the amount of human effort required to conduct systematic reviews~\cite{tsafnat2014systematic,marshall2019toward}, 
but the methods developed to support the screening task are trained on data from a small number of systematic reviews and are not yet able to fully replace humans~\cite{o2015using}.

An alternative is to find ways to map all clinical trials to standardised representations of the populations, 
interventions, comparators, and outcomes~(PICO) and aggregate information across studies that answer equivalent clinical questions. 
PICO extraction is a well-studied problem structure and was used as one of the examples in the development of SciBERT~\cite{DBLP:conf/emnlp/BeltagyLC19}.

Many PICO extraction methods focus on annotating sentences that include PICO information. 
However, if the goal is to fully automate a process for augmenting a systematic review with new studies as their results become available, 
even a perfect annotation of just the sentence is not enough. 
An expert will still need to read the sentence and then extract and normalise the information representing the population, 
intervention, comparator, and primary outcomes~(PICO) from those sentences. 
Full automation of the task requires the ability to identify the text spans that represent the PICO information.

Named entity recognition~(NER) seeks to identify the types and boundaries of targeted spans in unstructured text data. 
Machine learning NER methods have been based on BiLSTM-CRF~\cite{DBLP:conf/naacl/LampleBSKD16} and BERT~\cite{DBLP:conf/naacl/DevlinCLT19}, 
and these require human annotated data for training. 
For other application domains where span annotation is designed for entities such as people, locations, and organisations, 
crowdsourcing of labels is somewhat easier because a broader range of people can annotate data without specialised training. 
Even when annotated by domain experts, there is still substantial inconsistency across annotators~\cite{DBLP:conf/sigir/LeeS19}, 
though sentence level annotations tend to be more consistent~\cite{DBLP:conf/emnlp/ZlabingerSHH20}.

There is a gap in both the volume of available training data and demonstrated performance between NER in general domains such as news and biomedical applications compared to PICO extraction. 
There is a clear need for new approaches that can handle the more complicated and challenging token structure of biomedical entities 
including unusual synonyms and subordinate clauses that might include numerical information~(\textit{e.g.} drug dosage, or test result threshold); 
and can incorporate domain-specific knowledge in intelligent and useful ways.

In this paper, we address these challenges by proposing a novel PICO annotation task design. 
We propose a simplified requirement for annotation where we only need to know whether a sentence includes any PICO information. 
We use this coarser set of annotations to learn and infer PICO sentence types and span detections.

The pre-trained neural language model~\cite{DBLP:conf/naacl/DevlinCLT19,DBLP:journals/bioinformatics/LeeYKKKSK20,DBLP:conf/bionlp/PengYL19} is firstly used as feature representation learning model for sentence classification to be fine-tuned, and identify whether a given sentence contains PICO spans. 
This makes the proposed approach also capable of inferring PICO sentences even without crowdsourced annotations.

To get the PICO spans, we apply a masked span prediction task to assist the inference process. 
The fine-tuned language model is then used as the task-specific knowledge provider for the PICO spans. 
Scored spans are then fed into an inference algorithm to produce the final detection results.

The contributions of this paper are as follows:
\begin{itemize}
\item We propose a span detection approach for PICO extraction that uses only low-quality, crowd-sourced, sentence-level annotations as inputs, which reduces the need for time-consuming annotations from experts.
\item We evaluate a novel structure for identifying candidate PICO sentences and masked span inference together. 
The masked span inference task replaces input spans with pre-defined mask tokens and the language model is used to infer which spans contribute most to the PICO sentence classification results.
\item We demonstrate results that substantially improve on recall in span detection to align with the use case in the systematic review application domain on two benchmark datasets.
\end{itemize}

\section{Related Work}

\subsection{Neural Language Models}
Pre-trained deep neural language models, such as ELMo~\cite{DBLP:conf/naacl/PetersNIGCLZ18}, GPT~\cite{gpt}, BERT~\cite{DBLP:journals/bioinformatics/LeeYKKKSK20} and its variants~\cite{DBLP:journals/corr/abs-1907-11692,DBLP:conf/iclr/LanCGGSS20},
have brought significant performance improvements in a wide range of NLP tasks, such as relation extraction~\cite{DBLP:conf/acl/AltHH19}, entity resolution~\cite{DBLP:conf/aaai/LiMWSW21}, and question answering~\cite{DBLP:journals/bioinformatics/LeeYKKKSK20}.
These language models generally benefit from large scale text corpora.
A major advantage of these methods come from the way long dependency token relations are captured to produce contextualised representations.

To adapt language models for use in the biomedical domain, researchers took pre-trained language models and re-trained them with domain-specific corpora, including PubMed abstracts and full text articles. These were then applied to a diverse range of NLP tasks, such as named entity recognition~\cite{DBLP:journals/bioinformatics/LeeYKKKSK20,DBLP:conf/bionlp/PengYL19} and document classification~\cite{DBLP:conf/bionlp/PengYL19}. In this paper, we make use of pre-trained neural language models as the backbone model to learn task-oriented information.

\subsection{PICO Annotation and PICO Extraction}
There are multiple uses cases associated with representing clinical trial reports~(including article abstracts, registrations, and protocols) 
by the participant inclusion criteria including condition, the interventions used in each of the study arms, and the set of primary outcomes measured during the trial. 
Like named entity recognition~(NER), PICO detection aims to identify certain spans in the text corresponding to the each of the categories: population, interventions, and outcomes. 
While is it possible to apply general NER methods for the PICO extraction task~\cite{DBLP:conf/acl/NguyenWLNL17,DBLP:conf/acl/NenkovaLYMWNP18,DBLP:journals/jamia/KangPTTW21}, 
there are several key differences that make general NER methods less effective. 
These differences include spans that often do not have distinguishing features such as capitalised tokens and PICO elements that are not limited to noun phrases.

Most PICO extraction methods are fully supervised and need annotated data, 
which requires expertise and can be time consuming. 
To acquire enough PICO annotations for training, researchers have developed methods that use crowdsourcing as an alternative~\cite{DBLP:conf/acl/NguyenWLNL17,DBLP:conf/acl/NenkovaLYMWNP18}. 
This results in low-quality annotations, especially for the boundaries of the spans. 
To improve the annotation quality, \newcite{DBLP:conf/emnlp/ZlabingerSHH20} simplified the annotation task 
from document level to sentence level and additionally guided workers with similar sentences 
that had already been annotated using an unsupervised semantic short-text similarity method. 
In this paper, we instead only require sentence-level crowdsourced annotations without a boundary information 
developing novel method to predict PICO spans without training data.

\subsection{Span based Methods}
Compared with token based methods such as BiLSTM-CRF~\cite{DBLP:conf/naacl/LampleBSKD16,DBLP:conf/acl/MaH16},
span based methods treat the spans~(\textit{i.e.} consecutive tokens), as the targets.
In one stream of research advances in span based methods, the aim is to extract the hidden representations of each token with the raw token sequence as input, then either use boundary token representations~\cite{DBLP:conf/emnlp/OuchiS018,DBLP:conf/acl/EbnerXCRD20} or aggregate all the token representations~\cite{DBLP:conf/aaai/LiuSLWZ20} as the span representation.
All possible spans are enumerated, classified, and decoded.
An alternative stream of span-based research aims to mask a span in the token sequence and recover the masked tokens with hidden representations~\cite{DBLP:journals/tacl/JoshiCLWZL20}.
Both research streams use supervised or self-supervised methods and high-quality annotations as training data.
In this paper, we instead extract the information stored in the span using only sentence-level annotations derived from crowdsourcing.

\section{Method}
\label{sec:method}
In this section, 
we define the task and then introduce
the proposed approach with BLUE~\cite{DBLP:conf/bionlp/PengYL19}, a BERT~\cite{DBLP:conf/naacl/DevlinCLT19} structured neural language model in the biomedical domain, as our backbone model and the inference algorithm.

\subsection{Task Definition}
\label{sec:problem}
Generally, to construct a dataset for PICO span detection, the annotators are presented with the full text, \textit{i.e.}, the entire abstract of a clinical trial report~\cite{DBLP:conf/acl/NguyenWLNL17,DBLP:conf/acl/NenkovaLYMWNP18}.
To improve the quality of the annotation, ~\newcite{DBLP:conf/emnlp/ZlabingerSHH20} proposed a novel annotation task, 
asking annotators to annotate sentences instead of abstracts with retrieved expert annotated sentences as examples based on sentence similarity methods.
Both annotation tasks require annotators to locate the boundaries of PICO spans.

However, getting agreed boundaries for PICO span annotation is challenging, especially for crowdsourcing annotators in the biomedical domain.
To fuse the gap between PICO sentence prediction and PICO span detection, we formalise the annotated dataset and the task as follows.
We represent the dataset as $D$ with $|D|$ sentences and a sentence from the dataset as $s$ with sentence annotations from $|C|$ annotators.
We use $\langle i, j\rangle$ to denote the boundaries of a PICO span starting from the $i$-th token~(inclusive) and ending before the $j$-th token~(exclusive).
The task is then to train a model $M$ that is able to yield a PICO span $\langle i, j\rangle$ with the probability $P(i,j|M, s, C)$.

\subsection{Sentence Classification Training}
\label{sec:sentence}
\begin{figure*}[ht]
    \centering
    \includegraphics[width=\textwidth]{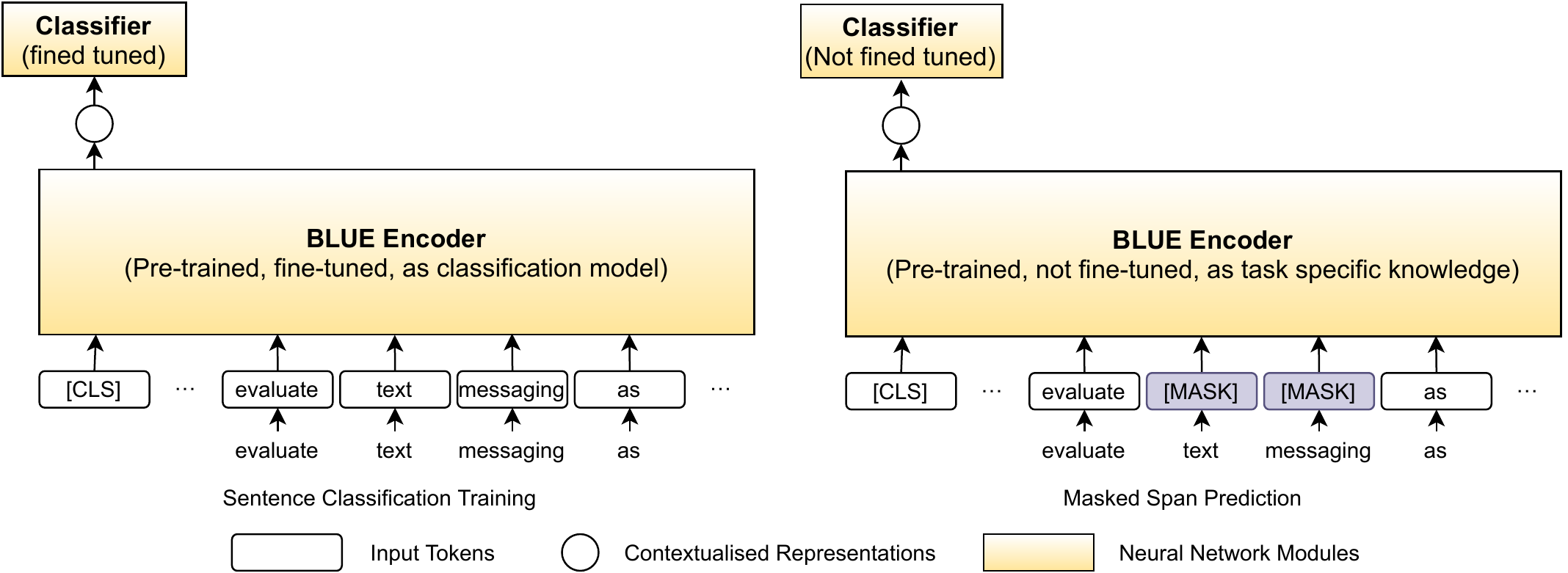}
    \caption{
    The proposed method has BLUE~\cite{DBLP:conf/bionlp/PengYL19} as the backbone model, 
    which is shared in the both the sentence classification~(the model on the left) and masked span prediction~(the model on the right).
    The BLUE encoder is firstly trained on the sentence prediction using the contextualised embedding of \texttt{[CLS]} token~(the left part of the figure).
    Then the fine-tuned BLUE encoder predicts the score with span masked~(the right part of the figure).
    The predicted scores of candidate spans along with the raw token sequence is collected for inference.
    }
    \label{fig:model}
\end{figure*}

The pre-trained neural language model BLUE is trained on domain specific text~(\textit{e.g.} PubMed abstracts). 
This injects the neural language model with biomedical knowledge.

From the annotators, we have gathered whether a sentence contains PICO annotations or not for each sentence.
These annotations represent task specific information and can be used to ``teach'' the model.
Thus, we form this process as a sentence classification task and fine-tune the pre-trained neural language model BLUE to predict whether a given sentence is a PICO sentence.

For each sentence, we feed it into the pre-trained BLUE model and collect the contextualised representations $\mathbf{h} \in \mathbb{R}^h$ of \texttt{[CLS]} token.
With this representation $\mathbf{h}$, we predict the probability of a PICO sentence using:
\begin{align}
score(y \mid \mathbf{h}) &= W\mathbf{h}+\mathbf{b},\\
p(y \mid \mathbf{h}) &= \text{Softmax}(score(y \mid \mathbf{h})),
  \label{eq:pre}
\end{align}
where $W \in \mathbb{R}^{h*h}$ and $\mathbf{b} \in \mathbb{R}^{h}$ are learnable parameters trained during the fine-tuning process, 
and $y$ is a function of $C$ annotations $y=f(C_1, \dots, C_{|C|})$~(details of the functions are presented in Section \ref{sec:data}).
Compared with other state-of-the-art text classification methods~\cite{DBLP:conf/emnlp/HuangMLZW19,DBLP:conf/acl/ZhangYCWWW20}, 
our sentence classification method is relatively simple with just one extended module.
The corresponding loss function we apply here is the cross entropy function:
\begin{equation}
\mathcal{L} = - \sum_{s \in D}ylog(p(y \mid \mathbf{h}).
\end{equation}

\subsection{Masked Span Prediction}
\label{sec:maskspan}
Like how human beings locate PICO spans given known PICO sentences, we want the model to focus on the most indicative token spans in the sentences.
A straightforward way is to directly use the last contextualised representations of the fine-tuned model to make predictions.
However, such an approach is problematic because: first, the last contextualised representations contain information from the corresponding tokens and the 
surrounding tokens. Thus taking either the span boundary representations or inner span representations would introduce a faint amount of interference 
and lead to error-prone results. Second, these representations show the information in the token level rather than span level, which does not treat the span as a whole.

To address the aforementioned problems, we apply a Masked Span Prediction~(MSP) task which is similar to the Masked Language Model training task applied in BERT~(shown in Fig~\ref{fig:model}). 
In the MSP task, for span $\langle i, j\rangle$, we firstly mask the original token sequence with \texttt{[MASK]} token.
This masked token sequence is put into the fine-tuned neural language model to get the prediction score $score_{\langle i, j\rangle}$.
The score $score_{\langle i, j\rangle}$ is then compared with the original score $score$ to infer the impact by the span $\langle i, j\rangle$, \textit{i.e.}, the contribution of the span in classifying the sentence as a PICO sentence. 
We define the contribution of a span $\langle i, j \rangle$ as $contribution_{\langle i, j\rangle} = score - score_{\langle i, j\rangle}$.
The contribution $contribution_{\langle i, j\rangle}$ could be either positive  or negative in classifying the sentence as a PICO sentence.

Given a sentence with $N$ tokens, there are $O(N^2)$ candidate spans to be masked, which could generate an intractable number of spans.
To reduce the number of candidate spans, we follow~\newcite{DBLP:conf/emnlp/OuchiS018} and~\newcite{DBLP:conf/acl/EbnerXCRD20} by limiting the number of tokens in a span up to $M$ tokens.
This would reduce the number of candidate spans to $\frac{M(2N-M+1)}{2}$, which is linear in the length of the sentence.

\begin{algorithm}[htbp]
\SetAlgoLined
\SetKwFor{ForEach}{for each}{do}{end for}%
\SetKwInOut{Input}{Input}
\SetKwInOut{Output}{Output}
\SetKwProg{Fn}{Function}{:}{}

\small%
\Input{Eliminated Span Set $RM$, Span $\langle i, j\rangle$} 
\Output{\textit{Updated} Eliminated Span Set $RM$, Span Remove Flag $F$}

$F \gets False$;

\For{$p \gets i+1$ \KwTo $j-1$}{\label{alg:sr1}
	\If{$\langle i, p\rangle \in RM \land \langle p, j\rangle \in RM$}{\label{alg:sr2}
		$RM \gets RM \cup {\langle i, j\rangle}$;\label{alg:sr3}

		$F \gets True$\;\label{alg:sr4}

		break\;
	}
}
\Return{$RM$, $F$}
\caption{Nested Span Elimination}
\label{alg:spanremove}
\end{algorithm}

We observe that some spans with a negative contribution to PICO sentence classification have at least one nested span split where the split spans have negative contributions as well.
To reduce the number of candidate spans for inference, we make use of this observation and apply with a bottom-up nested span elimination algorithm~(Algorithm~\ref{alg:spanremove}).
Eliminated Span Set $RM$ is initialised with single token spans of negative contributions to PICO sentence classification.
For spans with more than two tokens, we search every split of the target span~(Line~\ref{alg:sr1}).
When the split spans are both in the existing eliminated span set $RM$~(Line~\ref{alg:sr2}), 
we update $RM$ with the target span~(Line~\ref{alg:sr3}), 
claim this span can be eliminated~(Line~\ref{alg:sr4}), and return the result.
As we remove the spans based on model related set initialisation $RM$,
we report the percentage of eliminated candidate span in Section~\ref{sec:exp_spanreduc}. 

\subsection{Span Inference}
\label{sec:inference}

\begin{algorithm}[htbp]
\SetKwFor{ForEach}{for each}{do}{end for}%
\SetKwInOut{Input}{Input}
\SetKwInOut{Output}{Output}

\small%
\Input{Candidate Span Set $U$, the max select number $K$} 
\Output{Selective Span Set $R$}

$R \gets \emptyset$;

\tcc{descending sort $U$ by contribution}
$U \gets sorted(U)$;\label{alg:infer1}

\ForEach{$span \in U$}{\label{alg:infer2}
\tcc{$noneoverlap(R, span)$: whether $span$ has overlapped tokens with any spans in the $R$}
\If{$noneoverlap(R, span)$}{\label{alg:infer3}
	$R \gets R \cup {span}$;
	}
\If{$size(R) == K$}{\label{alg:infer4}
	break;
	}
}\label{alg:infer5}
\Return{$R$}
	\caption{Top-$K$ Inference}
	\label{alg:inference}
\end{algorithm}

With all the reserved candidate spans and their contributions, we need to select spans that (1) have positive contributions to PICO sentence classification and (2) are not nested spans.
Following previous work~\cite{DBLP:conf/emnlp/OuchiS018,DBLP:conf/acl/EbnerXCRD20}, we establish a similar \textit{argmax} inference method.
However, different from the tasks in~\newcite{DBLP:conf/emnlp/OuchiS018} and~\newcite{DBLP:conf/acl/EbnerXCRD20}, where each role is generally satisfied by exactly one span, 
there could be multiple spans appearing in one sentence in PICO span detection task.
Thus we apply a top-$K$ \textit{argmax} inference method with pre-defined $K$ for each PICO type.
As shown in Algorithm~\ref{alg:inference}, we firstly sort the candidate span set by span contribution~(Line~\ref{alg:infer1}), and iteratively select 
spans~(Line~\ref{alg:infer2}-\ref{alg:infer5}). If the span does not overlap with selected spans~(Line~\ref{alg:infer3}), we add it to the result set.
The algorithm ends when we get $K$ spans~(Line~\ref{alg:infer4}) or traverse all the reserved candidate spans.

\section{Experiments}
\label{sec:exps}
In this section, we evaluate our method and compare it with supervised, semi-supervised methods, and crowdsourced annotations on two benchmark datasets.
We also investigate the PICO sentence prediction results and the effect of candidate span reduction in the proposed method.
\subsection{Dataset}
\label{sec:data}

We use two benchmark datasets for the PICO span detection task:
EBM-NLP~\cite{DBLP:conf/acl/NenkovaLYMWNP18} dataset\footnote{\url{https://ebm-nlp.herokuapp.com/}}\footnote{We use the latest version 2.0 in this paper.} and 
PICO-data~\cite{DBLP:conf/acl/NguyenWLNL17} dataset\footnote{\url{https://github.com/yinfeiy/PICO-data}}.
The dataset statistics are shown in Table~\ref{tbl:data}. All datasets are in English language.

The PICO dataset includes 3549, 500, and 191 abstracts for training, development, and test sets, respectively.
This dataset is only annotated with Population~(P) type.

\begin{table}[htbp]
\centering
\caption{Dataset Statistic}
\resizebox{1\columnwidth}{!}{
\begin{tabular}{ccccc}
\toprule
\multirow{2}{*}{Dataset}&\multirow{2}{*}{PICO Type}&\multicolumn{3}{c}{\#Doc}\\
\cmidrule{3-5}
 & &Train&Dev&Test\\
\midrule
PICO-data~\cite{DBLP:conf/acl/NguyenWLNL17}&Population&3549&500&191\\
\midrule
\multirow{3}{*}{EBM-NLP~\cite{DBLP:conf/acl/NenkovaLYMWNP18}}&Population&4282&500&188\\
&Intervention&4282&500&189\\
&Outcome&4170&500&190\\
\bottomrule
\end{tabular}
}
\label{tbl:data}
\end{table}

\begin{table*}
\centering
\caption{Precision, Recall, and $F_1$ score for PICO Span Detection on the EBM-NLP dataset. 
Supervised and Semi-supervised methods are trained on the entire training set using the aggregated crowdsourced PICO span annotations. 
Weakly-supervised methods are trained on the entire training set \textit{only} using PICO sentence annotations. 
Human annotation are the aggregated crowdsourced PICO span annotations on the test set.
All methods are evaluated against expert span annotations on the test set.
We highlight the recall as it is the most important metric when PICO detection is applied for systematic review processes.
* indicates the results applying the crowdsourced sentence annotations.
}
\resizebox{2.1\columnwidth}{!}{
\begin{tabular}{llccccccccc}
\toprule
\multirow{2}{*}{Supervision Type}&\multirow{2}{*}{Method} & \multicolumn{3}{c}{Population} & \multicolumn{3}{c}{Intervention} & \multicolumn{3}{c}{Outcome}\\
\cmidrule{3-11}
&  & Precision    & Recall    & $F_1$    &  Precision& Recall    &  $F_1$     &  Precision& Recall    &  $F_1$     \\
\midrule
Human Annotation     &    HMMCrowd\cite{DBLP:conf/acl/NguyenWLNL17}    & 0.72 &0.76 &0.70&0.64 & 0.80 &0.68&0.50 &0.81 &0.59\\
\midrule
Supervised     &  CRF\cite{DBLP:conf/icml/LaffertyMP01}  &    0.55 & 0.51 &0.53&0.65 & 0.21 & 0.32&0.83 &0.17 &0.29\\
\midrule
Semi-Supervised     &  BiLSTM-CRF\cite{DBLP:conf/naacl/LampleBSKD16}      &0.78 &0.66 &0.71&0.61 &\textbf{0.70} &0.65&0.69 &0.58 &0.63\\
\midrule
\midrule
\multirow{3}{*}{Weakly-Supervised}     &  $\text{Sent2Span}_{agg}$ &0.31&0.78&0.44&0.22&0.50&0.31&0.30&0.55&0.39\\
      &  $\text{Sent2Span}_{major}$      &0.39&0.46&0.42&0.27&0.18&0.21&0.30&0.10&0.15\\
      &  $\text{Sent2Span}_{minor}$      &0.30&\textbf{0.85}&0.45&0.23&0.51&0.31&0.27&\textbf{0.64}&0.39\\
\midrule
\multirow{3}{*}{Weakly-Supervised*}     &  $\text{Sent2Span}_{agg}$ &0.32&0.79&0.46&0.22&0.54&0.31&0.31&0.57&0.40\\
      &  $\text{Sent2Span}_{major}$      &0.38&0.52&0.44&0.27&0.15&0.19&0.35&0.08&0.12\\
      &  $\text{Sent2Span}_{minor}$      &0.30&\textbf{0.86}&0.45&0.23&0.51&0.31&0.28&\textbf{0.64}&0.39\\
\bottomrule
\end{tabular}
}
\label{tbl:ebm}
\end{table*}

\begin{table}[ht]
\large
\centering
\caption{Precision, Recall, and $F_1$ score for PICO Span Detection on the PICO-data dataset. * indicates the results applying the crowdsourced sentence annotation.}
\resizebox{1\columnwidth}{!}{
\begin{tabular}{llccc}
\toprule
\multirow{2}{*}{Supervision Type}&\multirow{2}{*}{Method} & \multicolumn{3}{c}{Population}\\
\cmidrule{3-5}
 &  & Precision    & Recall    & $F_1$    \\
\midrule
Human Annotation     &    HMMCrowd   &0.73&0.75&0.74 \\
\midrule
Supervised     &  CRF  &0.80&0.55&0.65\\
\midrule
Semi-Supervised     &  BiLSTM-CRF      &0.74&0.65&0.69\\
\midrule
\midrule
\multirow{3}{*}{Weakly-Supervised}&$\text{Sent2Span}_{agg}$&0.32&0.72&0.45\\
&$\text{Sent2Span}_{major}$&0.39&0.52&0.45\\
&$\text{Sent2Span}_{minor}$&0.29&\textbf{0.87}&0.43\\
\midrule
\multirow{3}{*}{Weakly-Supervised*}&$\text{Sent2Span}_{agg}$&0.33&0.70&0.45\\
&$\text{Sent2Span}_{major}$&0.38&0.49&0.43\\
&$\text{Sent2Span}_{minor}$&0.29&\textbf{0.83}&0.44\\

\bottomrule
\end{tabular}
}
\label{tbl:pico}
\end{table}

\begin{table*}
\centering
\caption{Accuracy, Precision, Recall, and $F_1$ score of PICO sentence classification for crowdsourced annotations and proposed methods on the EBM-NLP dataset. 
$\text{Crowd}_{X}$ refers to the aggregated crowdsourced annotations mentioned in Section~\ref{sec:data}.
$\text{Sent2Span}_{X}$ refers to proposed methods trained with different PICO sentence annotations mentioned in Section~\ref{sec:data}.
The \textbf{bold-faced} scores represent the best results among crowdsourced annotations and proposed methods.
}
\resizebox{2.1\columnwidth}{!}{
\begin{tabular}{lcccccccccccc}
\toprule
\multirow{2}{*}{Method} & \multicolumn{4}{c}{Population} & \multicolumn{4}{c}{Intervention} & \multicolumn{4}{c}{Outcome}\\
\cmidrule{2-13}
  & Accuracy & Precision    & Recall    & $F_1$    & Accuracy &  Precision& Recall    &  $F_1$     & Accuracy &  Precision& Recall    &  $F_1$     \\
\midrule
$\text{Crowd}_{agg}$ 		&0.90&0.76&0.84&0.80&\textbf{0.90}&0.88&0.88&\textbf{0.88}&\textbf{0.84}&0.90&0.77&\textbf{0.83}\\
$\text{Crowd}_{major}$      &0.88&0.90&0.57&0.70&0.70&\textbf{0.99}&0.32&0.48&0.61&\textbf{0.96}&0.22&0.36\\
$\text{Crowd}_{minor}$      &0.73&0.47&\textbf{0.98}&0.64&0.83&0.73&\textbf{0.98}&0.83&0.78&0.71&0.97&0.82\\
\midrule
$\text{Sent2Span}_{agg}$ &\textbf{0.92}&0.84&0.84&\textbf{0.84}&0.88&0.91&0.80&0.85&0.81&0.89&0.70&0.79\\
$\text{Sent2Span}_{major}$      &0.86&\textbf{0.91}&0.44&0.59&0.65&0.96&0.22&0.36&0.56&0.93&0.13&0.23\\
$\text{Sent2Span}_{minor}$      &0.82&0.58&0.96&0.72&0.83&0.73&\textbf{0.98}&0.83&0.78&0.70&\textbf{0.98}&0.82\\
\bottomrule
\end{tabular}
}
\label{tbl:ebm-sent}
\end{table*}

\begin{table}[th]
\large
\centering
\caption{Accuracy, Precision, Recall, and $F_1$ score of PICO sentence classification on the PICO-data dataset. }
\resizebox{1\columnwidth}{!}{
\begin{tabular}{lcccc}
\toprule
\multirow{2}{*}{Method} & \multicolumn{4}{c}{Population}\\
\cmidrule{2-5}
  & Accuracy & Precision    & Recall    & $F_1$    \\
\midrule
$\text{Crowd}_{agg}$ 		&0.88&0.75&0.77&0.76\\
$\text{Crowd}_{major}$      &0.86&\textbf{0.88}&0.51&0.64\\
$\text{Crowd}_{minor}$      &0.78&0.54&0.90&0.68\\
\midrule
$\text{Sent2Span}_{agg}$ &\textbf{0.90}&0.82&0.81&\textbf{0.81}\\
$\text{Sent2Span}_{major}$      &0.85&0.86&0.51&0.64\\
$\text{Sent2Span}_{minor}$     &0.82&0.60&\textbf{0.96}&0.74\\
\bottomrule
\end{tabular}
}
% 191715
\label{tbl:pico-sent}
\end{table}

The EBM-NLP dataset includes 4,993 PubMed abstracts annotated with Population~(P), Intervention~(I), and Outcome~(O), respectively.
Comparator~(C) and Intervention~(I) are combined together as Intervention~(I).
As each PICO type is annotated individually to avoid cognitive load, this makes the dataset contain three sub-datasets with a single PICO type.
Without standard train-development split, we leave 500 abstracts from the training set as the development for each PICO type.

For all the training, development, and test sets in both datasets, they all collect the crowdsourced annotations, and the aggregated annotations.
The test sets also have expert annotations from medical experts\footnote{We note that some documents have no annotations and exclude them.}.
We additionally supply PICO sentence annotations by aggregating the existing crowdsourced annotations, indicating whether the sentence has at least one PICO span:
\begin{itemize}
\item $agg$: indicates whether a sentence received PICO annotations based on the aggregated annotations, to simulate the weighted annotator scenario
\item $major$: indicates whether a sentence received PICO annotations from more than half annotators
\item $minor$: indicates whether a sentence received at least one PICO annotation
\end{itemize}
On both datasets, we use spaCy~\footnote{\url{https://spacy.io/}} for tokenization and sentence split.

\subsection{Hyperparameters}
\label{sec:hyper}
We use the BLUE~\cite{DBLP:conf/bionlp/PengYL19} model trained on PubMed abstracts as the backbone model\footnote{We also experimented with the one trained on PubMed abstracts and MIMIC-III and found similar results.} with the BERT-Base structure and uncased tokenisation.
The maximum sequence length is set to 512 tokens. The training batch size is 32 and the evaluation batch size is 64.
For PICO sentence classification, we use Adam~\cite{DBLP:journals/corr/KingmaB14} optimiser and set the peak learning rate to 2e-5.
We train the models for 5 epochs and select the models with the best $F_1$ score on the development set.

For PICO Masked Span Prediction, the maximum span lengths $M$ are set to be 20, 7, and 10 tokens~\footnote{This is on the raw string token sequence and this may vary after Bert tokenization.}, for Population, Intervention, and Outcome, respectively. 
% In this case, we cover more than 90\% PICO spans according to the statical information on the development set.
The number of selected candidate spans $K$ is set to 2 for all the PICO types.
 % as more than 95\% PICO sentences have one or two PICO spans based on the statical information on the development set.
The maximum span lengths and the number of selected candidate spans are set based on the statistical information of the aggregated crowdsourced PICO spans on the development sets to cover at least 90\% and 95\% PICO spans, respectively.
All the experiments are run on one NVIDIA V100 GPU. Our source code is available online\footnote{Our implementation is publicly available at \url{https://github.com/evidence-surveillance/sent2span}}.

\subsection{Comparison Methods and Evaluation Metrics}
\label{sec:compare}
Following previous work~\cite{DBLP:conf/acl/NguyenWLNL17,DBLP:conf/acl/NenkovaLYMWNP18}, we use the token-wise precision, recall, 
and $F_1$ score of the output PICO spans against the expert annotations. 
Following previous work~\cite{thomas2021machine}, we focus on recall as it is the most important metric when PICO detection is applied for systematic review processes.
We also report the accuracy, precision, recall and $F_1$ for the PICO sentence classification results against the expert annotations.

We compare Sent2Span against supervised method, semi-supervised method, and aggregated human annotations. All the methods are evaluated on the test set of expert annotations.

\begin{itemize}
\item HMMCrowd~\cite{DBLP:conf/acl/NguyenWLNL17}: HMMCrowd extends Dawid-Skene model~\cite{dawid1979maximum} with a HMM component, and explicitly uses the sequential structure of spans. This model is directly applied on the crowdsourced annotations to get the aggregated annotations without training.

\item Conditional Random Fields~(CRF)~\cite{DBLP:conf/icml/LaffertyMP01}: The CRF model is fully supervised with a feature template, including the current, previous, and next words; part-of-speech tags; and character information such as whether a token contains digits, uppercase letters, symbols, etc. This model is trained with the aggregated crowdsourced span annotations on the training dataset.

\item BiLSTM-CRF~\cite{DBLP:conf/naacl/LampleBSKD16}: BiLSTM-CRF model is a semi-supervised method with pre-trained word2vec embeddings trained on PubMed abstracts. This model is trained with the aggregated crowdsourced span annotations on the training dataset.

\item $\text{Sent2Span}_{X}$: This is our proposed method with different sentence annotation generation functions $X$. 
$X$ refers to the $agg$, $major$, and $minor$ mentioned in Section~\ref{sec:data}.
This method is trained only with sentence annotations without any span annotations.
This method is evaluated on the test datasets using both the predicted PICO sentence classification results and the crowdsourced PICO sentence annotations.
\end{itemize}

\subsection{PICO Span Detection Results}
\label{sec:results}
Table~\ref{tbl:ebm} and Table~\ref{tbl:pico} show the performance of different methods on the EBM-NLP and PICO-data datasets.
$\text{Sent2Span}_{minor}$ always shows the best recall with the best $F_1$ score in most case among the rest Sent2Span models.
Compared with the rest methods, $\text{Sent2Span}_{minor}$ achieves the best recall for the Population and Outcome types, 
and even better than the aggregated human annotation results for Population.

It surpasses the aggregated human annotation by 10\% recall on Population on both datasets, 
even though the crowdsourcing annotators have been given annotation guidelines and certain examples.
For Outcome, it achieves better recall compared to supervised and semi-supervised methods by around 6\%, 
while these methods are trained with span-level annotations compared with the sentence-level annotation applied for Sent2Span.
Though Sent2Span does not achieve the best recall for Intervention, it beats the supervised CRF methods by a large margin~(0.51 vs 0.21).

Sent2Span does not exhibit a performance drop when applying sentence-level human annotation for inference.
It shows better recall~(\textit{i.e.} Population on the PICO dataset) for all the proposed models.
This indicates: (1) Sent2Span generalises well on the datasets; and (2) Sent2Span can be directly applied for PICO span detection.

\subsection{PICO Sentence Classification Results}
\label{sec:sent_class}

As the results are inferred from the PICO sentence classification results, it is worth examining the corresponding impact.
We show the results of crowdsourced annotation and Sent2Span in Table~\ref{tbl:ebm-sent} and Table~\ref{tbl:pico-sent} for both datasets.

From both tables, $agg$ results have the best accuracy and $F_1$ score, $major$ results have the best precision, and $minor$ results have the best recall.
$\text{Sent2Span}_{minor}$ shows the better recall for Intervention and Outcome on the EBM-NLP dataset, and Population on the PICO dataset compared with crowdsourced annotations.
This explains the equivalent and superior recall results in PICO span detection~(see Table~\ref{tbl:ebm} and Table~\ref{tbl:pico}).
Without losing potential PICO sentences,  $\text{Sent2Span}_{minor}$ selects the most PICO spans.
And $\text{Sent2Span}_{major}$ has the best precision with the worst recall.

\subsection{Effect of Candidate Span Reduction}
\label{sec:exp_spanreduc}

\begin{table}[th]
\large
\centering
\caption{Candidate Span Reduction with Algorithm~\ref{alg:spanremove}. The number indicates the proportion of reduced candidate spans against all the candidate spans.}
\resizebox{1\columnwidth}{!}{
\begin{tabular}{lcccc}
\toprule
\multirow{2}{*}{Method} & PICO& \multicolumn{3}{c}{EBM-NLP}\\
\cmidrule{2-5}
  & Population & Population    & Intervention    & Outcome    \\
\midrule
$\text{Sent2Span}_{agg}$ &0.15&0.20&0.16&0.21\\
$\text{Sent2Span}_{major}$      &0.22&0.37&0.20&0.23\\
$\text{Sent2Span}_{minor}$      &0.08&0.15&0.12&0.08\\
\bottomrule
\end{tabular}
}
\label{tbl:span-reduction}
\end{table}

To demonstrate the effectiveness of the Nested Span Elimination Algorithm~\ref{alg:spanremove}\footnote{We see no major difference in performance when we do not reduce candidate spans (less than 0.002 in precision, recall, and F1-score) compared with the performance using the algorithm.} in the Masked Span Prediction task,
we count both the number of candidate spans with and without using the algorithm, and calculate the percentage of reduced candidate spans.
The results are listed in Table~\ref{tbl:span-reduction} for both datasets.

At least 8\% of candidate spans are eliminated and do not pass to the fine-tuned neural language model for inference, which saves inference time.
In Comparison to models trained with different sentence-level annotations, 
$\text{Sent2Span}_{major}$ has the most discarded candidate spans with more than 22\% candidate spans eliminated cross both datasets.
And $\text{Sent2Span}_{minor}$ receives the least number of eliminated spans.
This indicates that $\text{Sent2Span}_{major}$ tends to ignore more spans 
and $\text{Sent2Span}_{minor}$ builds more connections between the spans and PICO sentence classification results resulting in higher recall.

\subsection{Error Analysis}

We perform an error analysis using the test datasets for which there are expert annotations.
Error types include boundary errors~(BE), overlap errors~(OE), false-positive errors~(FP), 
and false-negative errors~(FN)~(Table~\ref{tbl:error-example}).

\begin{table}
\centering
\caption{Five examples of predicted PICO spans (Prediction) and their error types~(boundary errors, overlap errors, false-positive errors, and false-negative errors). 
The illustration snippet contains 7 tokens, and tokens in blue are either annotated by experts or predicted by the methods (we assume the annotations and predictions have the same PICO type).
FP and FN indicate false-positive and false-negative errors, respectively. }
% o/x indicates a correct/wrong prediction against gold annotation (Gold) by a given agreement.}

\begin{tabular}{cccccc}
\toprule
\textbf{No.} &\textbf{Prediction}&\textbf{Gold}&  \textbf{Error Type}\\
\midrule
1&\includegraphics[width=0.1\textwidth]{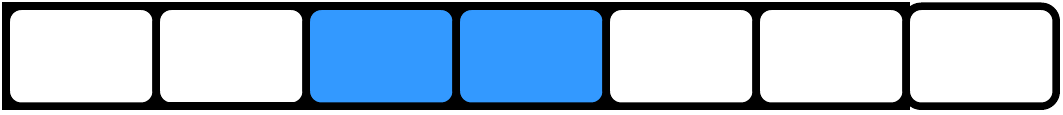} &\multirow{5}{*}{\includegraphics[width=0.1\textwidth]{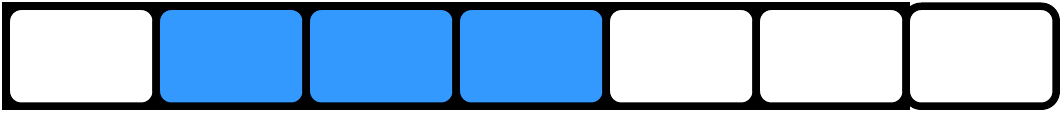}}& Boundary \\
2&\includegraphics[width=0.1\textwidth]{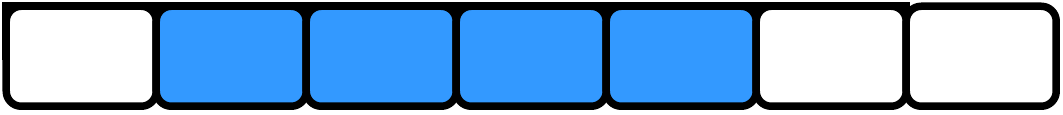} & & Boundary\\
3&\includegraphics[width=0.1\textwidth]{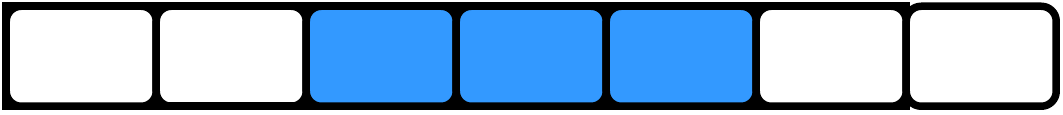} & & Overlap\\
4&\includegraphics[width=0.1\textwidth]{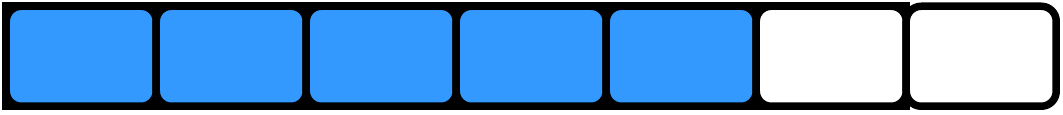} & & Overlap\\
5&\includegraphics[width=0.1\textwidth]{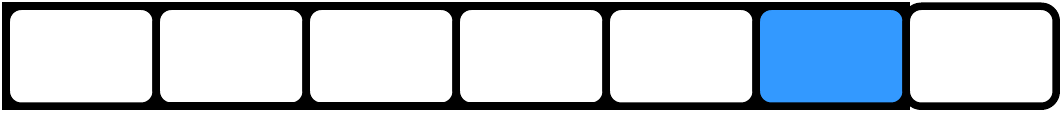} & & FP and FN\\
\bottomrule
\end{tabular}
\label{tbl:error-example}
\end{table}

\begin{table*}
\centering
\caption{The number of errors from the proposed methods on the EBL-NLP and PICO-data datasets. 
Crowd indicates the aggregated crowdsourced annotations.
BE indicates boundary error. OE indicates overlap error. FP and FN indicate false-positive and false-negative errors, respectively.}
\resizebox{2.1\columnwidth}{!}{
\begin{tabular}{lcccccccccccccccc}
\toprule
\multirow{3}{*}{Method} & \multicolumn{12}{c}{EBM-NLP} & \multicolumn{4}{c}{PICO-data}\\
\cmidrule{2-17}
	& \multicolumn{4}{c}{Population} & \multicolumn{4}{c}{Intervention} & \multicolumn{4}{c}{Outcome} & \multicolumn{4}{c}{Population}\\
\cmidrule{2-17}
	& BE & OE    & FP    & FN & BE & OE    & FP    & FN & BE & OE    & FP    & FN & BE & OE    & FP    & FN    \\
\midrule
Crowd & 161 & 10 & 185 & 109 & 314 & 47 & 303 & 237 & 408 & 22 & 235 & 542 & 178 & 29 & 203 & 172 \\
\midrule
$\text{Sent2Span}_{agg}$ & 207 & 354 & 467 & 112 & 356 & 524 & 574 & 650 & 342 & 659 & 650 & 692 & 170 & 418 & 526 & 154\\
$\text{Sent2Span}_{major}$ & 100 & 198 & 182 & 366 & 111 & 152 & 168 & 1398 & 68 & 130 & 147 & 1590 & 121 & 295 & 324 & 355\\
$\text{Sent2Span}_{minor}$ & 201 & 415 & 1032 & 45 & 357 & 561 & 891 & 641 & 443 & 844 & 1532 & 422 & 274 & 486 & 1125 & 42\\
\bottomrule
\end{tabular}
\label{tbl:error}
}
\end{table*}

% To further explore the limitations of the proposed methods, we establish a brief error analysis on the test datasets 
% as only the test datasets provide with expert annotations.
% % We establish the error anlysis on the test datasets 
% % We analysis the prediction results of the proposed methods and the crowdsourced annotations against the expert annotations on the test datasets 
% % as only the test datasets provide with expert annotations.
% We divide the errors into four types, including boundary errors~(BE), overlap errors~(OE), false-positive errors~(FP), 
% and false-negative errors~(FN) as illustrated in Table~\ref{tbl:error-example}.

The number of errors varies by type when comparing the proposed methods and the crowdsourced annotations~(Table~\ref{tbl:error}).
$\text{Sent2Span}_{major}$ gives the least number of BEs compared with the two other methods, at the cost of the largest number of FN errors and thus the lowest recall.
All proposed methods have more OEs when compared to crowdsourced annotations, suggesting an area where future work would be valuable.
In a post-hoc analysis of the OEs, we find that there are no nested predictions~(for example, ``training'' is nested in ``progressive muscle relexation training'').

$\text{Sent2Span}_{minor}$ produces the most FP errors and the least FN errors across all datasets, and the difference is most pronounced for the Population type. Overall, $\text{Sent2Span}_{minor}$ produces a high recall, low precision tool, and this is related to the way K (the number of selected candidate spans) is set, introducing redundant selected spans in the final inference result. This approach is likely to be useful as part of certain pipelines where PICO extraction is used, but there is also room for further improvement in the span selection mechanism.

% In Table~\ref{tbl:error}, we list the number of errors in different types for the proposed methods and the crowd annotations. 
% % $\text{Sent2Span}_{agg}$ generally has the least number of boundary errors compared with $\text{Sent2Span}_{major}$ and $\text{Sent2Span}_{minor}$ on two datasets.
% $\text{Sent2Span}_{major}$ represents the least number of boundary errors compared with the other two methods.
% However, it also shows the most number of false-negative errors explaining the worst recall performance compared with the other two methods.
% All the proposed methods have more overlap errors compared with crowdsourced annotations, which is the improvement direction in our future work.
% We examine all the overlap errors in the proposed methods and find there is no nested prediction~(For example, ``training'' is nested in ``progressive muscle relexation training'').
% $\text{Sent2Span}_{minor}$ produces the most false-positive errors and the least false-negative errors across all the datasets, especially for the Participant type.
% This is also reflected in the performance of $\text{Sent2Span}_{minor}$~(high recall and low precision).
% This is in connection to the setting of the number of selected candidate spans $K$, 
% which introduces redundant selected spans in the final inference result.
% We will further improve the inference method with an adapted span selection mechanism in furture work.

\section{Discussion}
\label{sec:discussions}

The results of the experiments show that our proposed method could be a useful component of a pipeline for PICO detection and extraction, 
in use cases where costly expert annotation is limited and where the aim is to identify all relevant examples. 
The results show the approach compares favourably to existing supervised methods 
and achieves high recall for PICO span detection even without using any span annotations as training data.

Our proposed approach is likely to be useful in a range of other application domains. 
Many NLP tasks across NER~\cite{DBLP:conf/emnlp/SohrabM18,DBLP:conf/aaai/LiuSLWZ20,DBLP:conf/acl/XiaZYLDWFMY19} and semantic role labelling~\cite{DBLP:conf/emnlp/OuchiS018} can be formulated as a span detection task. 
In cases where it is easier to acquire or estimate low-quality sentence-level annotations, 
and resource-intensive to acquire high-quality span-level annotations, our proposed approach may be appropriate.

Several opportunities exist for future work. 
For the sake of simplicity, we used the BLUE model~\cite{DBLP:conf/bionlp/PengYL19} with the BERT-base-uncased structure, 
rather than exploring models with BERT-large-cased structure. 
We also did not explore the use of other NLP tools such as part-of-speech tagging or dependency parsing. 
Integration of these methods in the proposed~Sent2Span approach may improve the performance.

The~Sent2Span method is designed to be part of a larger pipeline of techniques designed to support systematic review processes. 
Other than PICO detection, extraction, and representation, other methods have been proposed for identifying which trials should be included in systematic review updates~\cite{DBLP:journals/jbi/SurianDOBCB18}. 
Future work in the space could include head-to-head comparisons of approaches for rapid systematic review updating that test for maximising the completeness of evidence identification and minimising human effort.

\section{Conclusion}

The~Sent2Span method for PICO span detection we propose and test in this paper could be used to support new tasks in systematic review processes. 
The difference between~Sent2Span and previous approaches to PICO detection include 
the use of only low-quality sentence-level annotations as training data and 
the results demonstrating achieve high recall in span detection, which is an important requirement for systematic review processes.

\section*{Acknowledgements}
This work is supported by National Library of Medicine, National Institutes of Health under grant No.R01LM012976.

\bibliography{emnlp2021}
\bibliographystyle{acl_natbib}

\end{document}